\documentclass[runningheads]{llncs}
\pdfoutput=1
\usepackage{graphicx}
\usepackage{booktabs}
\usepackage{makecell}
\usepackage{multirow}
\usepackage{xspace}
\usepackage{enumitem}
\usepackage{todonotes}
\usepackage{microtype}
\usepackage{url}
\usepackage{hyperref}\usepackage[absolute]{textpos}

\newcommand{\headingNoSkip}[1]{\noindent\textbf{#1}\xspace}
\newcommand{\headingWithSkip}[1]{\smallskip\headingNoSkip{#1}}
\makeatletter\newcommand{\heading}{\@ifstar\headingNoSkip\headingWithSkip}\makeatother

\newcommand{\lindaturl}{\url{https://hdl.handle.net/11234/1-3691}\xspace}
\newcommand{\hfurl}{\url{https://huggingface.co/ufal/robeczech-base}\xspace}

\newenvironment{citemize}{\begin{itemize}[noitemsep,topsep=0pt]}{\end{itemize}}

\begin{document}
\begin{textblock}{16}(0,0.1)\centerline{\small This paper was published in \textbf{TSD 2021} -- please cite the published version \url{https://doi.org/10.1007/978-3-030-83527-9\_17} instead.}\end{textblock}
%
\title{RobeCzech: Czech RoBERTa, a monolingual contextualized language representation model}
\titlerunning{RobeCzech: Czech RoBERTa}
%
\author{
Milan Straka\orcidID{0000-0003-3295-5576} \and
Jakub Náplava\orcidID{0000-0003-2259-1377} \and\\
Jana Straková\orcidID{0000-0003-0075-2408} \and
David Samuel\orcidID{0000-0003-2866-1022}
}

\authorrunning{M. Straka et al.}
%
\institute{Charles University, Faculty of Mathematics and Physics, Institute of Formal and Applied Linguistics, Malostranské nám. 25, 118 00 Prague, Czech Republic
\email{\char`\{straka,naplava,strakova,samuel\char`\}@ufal.mff.cuni.cz}
}

\maketitle              

\begin{abstract}
We present RobeCzech, a monolingual RoBERTa language representation model trained on Czech data. RoBERTa is a robustly optimized Transformer-based pretraining approach. We show that RobeCzech considerably outperforms equally-sized multilingual and Czech-trained 
contextualized language representation models, surpasses current state of the art in all five evaluated NLP tasks and reaches state-of-the-art results in four of them. The RobeCzech model is released publicly at \lindaturl and \hfurl.

\end{abstract}


\section{Introduction}

We introduce RobeCzech: Czech RoBERTa, a Czech contextualized language representation model based on the Transformer architecture and trained solely on Czech data. RobeCzech is a monolingual version of RoBERTa \cite{RoBERTa}, a robustly optimized BERT \cite{devlin2018bert} pretraining approach.

In this paper, we describe the RobeCzech training process and we evaluate RobeCzech in comparison with current multilingual and Czech-trained contextualized language representation models: multilingual BERT \cite{devlin2018bert}, multilingual XLM-RoBERTa \cite{XLM-R} (base and large), Slavic BERT \cite{SlavicBERT} tuned on 4 Slavic languages, including Czech; and Czert \cite{sido2021czert}, another monolingual, Czech BERT model.
%

We show that RobeCzech considerably outperforms all models of similar size, and at the same time, it reaches new state-of-the-art results in four NLP tasks: morphological tagging and lemmatization, dependency parsing, named entity recognition and semantic parsing. In the last evaluated task, the sentiment analysis, RobeCzech also improves over state of the art and delivers the best results of all models of similar size, only being surpassed by XLM-RoBERTa large \cite{XLM-R}, a model 4 times the size of all the other evaluated models (Table~\ref{tab:n_params}).

We release the RobeCzech model for public use.

\section{Related Work}

Contextualized language representation models have recently accelerated progress in NLP. Significant advances have been reached particularly with Bidirectional Encoder Representations from Transformers, widely known as BERT \cite{devlin2018bert}, inspiring interest in Transformer-like architectures. We especially highlight RoBERTa \cite{RoBERTa} and its derivation XLM-RoBERTa \cite{XLM-R}.

The above mentioned language representation models were trained either only on English or as multilingual, though with an (implicit) strength in the most represented languages (i. e., English). Therefore, research has recently been focusing on monolingual BERT models, giving birth to national BERT mutations, e.g. French \cite{Camembert}, Finnish \cite{FinnishBERT}, Romanian \cite{RomanianBERT} and Czech \cite{sido2021czert}.

Our model is similar to the above mentioned Czert \cite{sido2021czert} in the sense that it is also a Czech contextualized language representation model, but unlike Czert, which is based on BERT, we trained a Czech version of RoBERTa. According to both the original Czert results \cite{sido2021czert} and the hereby presented evaluation on five NLP tasks, RobeCzech is better than Czert in all experiments by a considerable margin.

\section{Training the Czech RoBERTa}
\label{sec:training}

We trained RobeCzech on a collection of the following publicly available texts:
\begin{citemize}
    \item SYN v4 \cite{data:synv4}, a large corpus of contemporary written Czech, 4,188M tokens;
    \item Czes \cite{data:czes}, a collection of Czech newspaper and magazine articles, 432M tokens;
    \item documents with at least 400 tokens from the Czech part of the web corpus W2C \cite{data:w2c,majlis-w2c-tsd}, tokenized with MorphoDiTa~\cite{ACL2014}, 16M tokens;
    \item plain texts extracted from Czech Wikipedia dump 20201020 using WikiExtractor,\footnote{\url{https://github.com/attardi/wikiextractor}} tokenized with MorphoDiTa~\cite{ACL2014}, 123M tokens.
\end{citemize}

\noindent All these corpora contain whole documents, even if the SYN v4 is block-shuffled (blocks with at most 100 words respecting sentence boundaries are permuted in a document) and in total contain 4,917M tokens.

The texts are tokenized into subwords with a byte-level BPE (BBPE) tokenizer~\cite{radford2019language}. The tokenizer is trained on the entire corpus and we limit its vocabulary size to 52,000 items. 

\looseness-1
The RobeCzech model is trained using the official code released in the Fairseq library.\footnote{\url{https://github.com/pytorch/fairseq/blob/master/examples/roberta/}} The training batch size is 8,192 and each training batch consists of sentences sampled contiguously, even across document boundaries, such that the total length of each sample is at most 512 tokens (\textit{FULL-SENTENCES} setting~\cite{RoBERTa}). We use Adam optimizer \cite{Adam} with $\beta_1=0.9$ and $\beta_2=0.98$ to minimize the masked language-modeling objective. The learning rate is adapted using the polynomial decay schema with 10,000 warmup updates and the peak learning rate set to $7 \cdot 10^{-4}$. A total amount of 91,075 optimization steps were performed, which took approximately 3 months on 8 QUADRO P5000 GPU cards.  





\begin{table}
\caption{Number of parameters.}
\label{tab:n_params}
\centering
\newcommand{\0}{\hphantom{0}}
\setlength{\tabcolsep}{5pt}
\begin{tabular}{@{}lccc@{}}
\toprule
\textbf{Model} & \textbf{Embedding} & \textbf{Transformer} & \textbf{Total Parameters} \\ \midrule
mBERT uncased & \082M & \085M & 167M \\
Czert & \024M & \085M & 109M \\
Slavic BERT & \092M & \085M & 177M \\
XLM-R base & 192M & \085M & 277M \\
XLM-R large & 257M & 302M & 559M \\
\textbf{RobeCzech} & \040M & \085M & 125M \\ \bottomrule
\end{tabular}

\end{table}

\section{Evaluation Tasks}

We evaluate our Czech RoBERTa model on five NLP tasks in comparison with a variety of recently proposed mono- and multi-lingual contextualized language representation models
(to our best knowledge, these are all publicly available models trained at least partially on Czech):
\begin{citemize}
\item \textbf{mBERT \cite{devlin2018bert}:} well-known multilingual BERT language representation model.
\item \textbf{Czert \cite{sido2021czert}:} the first Czech monolingual model based on BERT.
\item \textbf{Slavic BERT \cite{SlavicBERT}:} multilingual BERT tuned specifically for NER on 4 Slavic languages data (Russian, Bulgarian, Czech and Polish).
\item\textbf{XLM-RoBERTa \cite{XLM-R}, base and large:} multilingual contextualized representations trained at large scale.
\end{citemize}

\noindent Except for XLM-RoBERTa large, which is 4 times larger than others, all models are of \textit{base} size \cite{devlin2018bert}, see Table~\ref{tab:n_params}.

We evaluate RobeCzech in five NLP tasks, three of them leveraging frozen contextualized word embeddings, two approached with fine-tuning:
\begin{citemize}
    \item \textbf{morphological analysis and lemmatization:} frozen contextualized word embeddings,
    \item \textbf{dependency parsing}: frozen contextualized word embeddings,
    \item \textbf{named entity recognition:} frozen contextualized word embeddings,
    \item \textbf{semantic parsing:} fine-tuned,
    \item \textbf{sentiment analysis:} fine-tuned.
\end{citemize}

\subsection{Morphological Tagging and Lemmatization on PDT 3.5}
\label{sec:tagging-pdt3.5}

\heading*{Dataset} We evaluate the morphological POS tagging and lemmatization on the morphological layer of the \textit{Prague Dependency Treebank 3.5} \cite{PDT3.5}.

\heading{Metric} The morphological POS tagging and lemmatization is evaluated using accuracy.

\heading{Architecture} We adopt the \textit{UDPipe~2} architecture \cite{UDPipe2.0}, reproducing the methodology of \cite{StrakaTSD2019}. After embedding input words, three bidirectional LSTM layers \cite{Hochreiter:1997:LSTM} are applied, followed by a softmax classification layer for POS and lemmas. In case of lemmas, the network predicts a simple edit script from input form to desired lemma. Since edit patterns are shared between lemmas due to regularities in morphology, the output categorization layer is reduced from the full vocabulary to only $1568$ classes (in PDT 3.5 \cite{PDT3.5}). In all our experiments, we use the same word embeddings as \cite{StrakaTSD2019}: pretrained \texttt{word2vec} embeddings \cite{mikolov}, end-to-end word embeddings and character-level word embeddings \cite{Cho2014,Graves2005,Ling2015}. The contextualized word embeddings are used frozen-style as additional inputs to the neural network.

\subsection{Dependency Parsing on PDT 3.5}
\label{sec:parsing-pdt3.5}

\heading*{Dataset} We evaluate the dependency parsing on the analytical layer of the \textit{Prague Dependency Treebank 3.5} \cite{PDT3.5}.

\heading{Metric} In evaluation, we compute both the unlabeled attachment score (UAS) and labeled attachment score (LAS).

\heading{Architecture} We perform dependency parsing jointly with POS tagging and lemmatization, following the experiments of \cite{StrakaTSD2019} showing that this approach is superior to using predicted POS tags and lemmas on input. We utilize the \textit{UDPipe~2} architecture \cite{StrakaTSD2019}: after embeddings input words and three bidirectional LSTM layers \cite{Hochreiter:1997:LSTM}, a biaffine attention layer \cite{dozat:2016} produces labeled dependency trees. The input word embeddings are the same as in the previous Section~\ref{sec:tagging-pdt3.5} and the contextualized word embeddings are additionally concatenated to the baseline input.

\subsection{Morphosyntactic Analysis on Universal Dependencies}

\heading*{Dataset} We further evaluate the joint morphosyntactic analysis on the \textit{UD Czech PDT} treebank of the \textit{Universal Dependencies 2.3} \cite{UD2.3-data}.

\heading{Metric} We use the standard evaluation script from \textit{CoNLL 2018 Shared Task: Multilingual Parsing from Raw Text to Universal Dependencies} \cite{CoNLL2018}, which produces the following metrics:
\begin{citemize}
    \item \textbf{UPOS} -- universal POS tags accuracy,
    \item \textbf{XPOS} -- language-specific POS tags accuracy,
    \item \textbf{UFeats} -- universal subset of morphological features accuracy,
    \item \textbf{Lemmas} -- lemmatization accuracy,
    \item \textbf{UAS} -- unlabeled attachment score, \textbf{LAS} -- labeled attachment score,
    \item \textbf{MLAS} -- morphology-aware LAS, \textbf{BLEX} -- bi-lexical dependency score.
\end{citemize}

\heading{Architecture} Following Section~\ref{sec:parsing-pdt3.5}, we employ the \textit{UDPipe~2}~\cite{StrakaTSD2019} architecture with frozen contextualized word embeddings.

\subsection{Named Entity Recognition}

\heading*{Dataset} We evaluate the Czech NER on all versions of the \textit{Czech Named Entity Corpus}, both the original \cite{CNEC} with nested entities and the CoNLL version \cite{CNECExtended} with reduction to flat entities only.

\heading{Metric} The standard evaluation metric for NER is F1 score computed over detected named entities spans.

\heading{Architecture} We reproduce the the current NER SoTA architecture \cite{StrakovaACL2019}, using the \textit{LSTM-CRF} and \textit{seq2seq} variants for flat and nested NER, respectively. All experiments include the Czech FastText word embeddings \cite{FastText} of dimension 300, end-to-end trained word embeddings and character-level word embeddings \cite{Cho2014,Graves2005,Ling2015} as inputs to the network. The contextualized word embeddings are used as frozen, additional inputs to the network.

\subsection{Semantic Parsing on Prague Tectogrammatical Graphs} 

\heading*{Dataset}
We use the \emph{Prague Tectogrammatical Graphs} (PTG) provided for the CoNLL 2020 shared task, \emph{Cross-Framework Meaning Representation Parsing} (MRP 2020) \cite{oepen-etal-2020-mrp}. The original annotation comes from the tectogrammatical layer of the \emph{Prague Dependency Treebank} \cite{PDT3.5};
the graphs for the shared task were obtained by relaxing its original limitation to trees – for example by explicitly modeling co-reference by additional edges instead of special node attributes \cite{zeman-hajic-2020-fgd}.

\heading{Metric}
We employ the official metric from MRP 2020, which first finds the maximum common edge subgraph to align the evaluated and the target graph. Then, it computes the micro-averaged F1 score over different features of the semantic graphs -- top nodes, node labels, node properties, anchors, edges between nodes and edge attributes. 

\heading{Architecture} We reimplement the current SoTA architecture for PTG parsing called \emph{PERIN} \cite{samuel-straka-2020-ufal}. This model does not assume any hard-coded ordering of the graph nodes, but instead dynamically finds the best matching between the predicted and the target ones.

Following \emph{UDify} \cite{udify}, we compute the contextualized subword embedding by taking the weighted sum of all hidden layers in a language representation model. The scalar weight for each layer is a learnable parameter. To obtain a single embedding for every token, we sum the embeddings of all its subwords. Finally, the summed embeddings are normalized with layer normalization \cite{layer_norm} to stabilize the training.

The pretrained encoder is finetuned with a lower learning rate than the rest of the model. The learning rate follows the inverse square root schedule with warmup and is frozen for the first 2000 steps before the warmup starts. The warmup phase takes 6000 steps and the learning rate peak is $6\cdot10^{-5}$.

\subsection{Sentiment Analysis}

\heading*{Dataset} We evaluate sentiment analysis on Czech Facebook dataset~(CFD)~\cite{data:czech_facebook,habernal-etal-2013-sentiment}. This dataset contains 2,587 positive, 5,174 neutral and 1,991 negative posts (the 248 bipolar posts are ignored, following \cite{habernal-etal-2013-sentiment,sido2021czert}).

\heading{Metric} The performance is evaluated using macro-averaged F1 score. Because the dataset has no designed test set, we follow the approach of the dataset authors~\cite{habernal-etal-2013-sentiment} and perform 10-fold cross-validation, reporting mean and standard deviation of the folds' F1 scores.

\heading{Architecture} We employ the standard text classification architecture consisting of a BERT encoder, followed by a softmax-activated classification layer processing the computed embedding of the given document text obtained from the \verb|CLS| token embedding from the last layer~\cite{devlin2018bert,RoBERTa}.%

We train the models using a lazy variant of the Adam optimizer~\cite{Adam} with a batch size of 64. During the first epoch, the BERT encoder is frozen and only the classifier is trained with the default learning rate of $10^{-3}$. From the second epoch, the whole model is updated, starting by 4 epochs of cosine warm-up from zero to a specified peak learning rate, followed by 10 epochs of cosine decay back to zero.

We consider peak learning rates $10^{-5}, 2\cdot10^{-5}, 3\cdot10^{-5}$ and $5\cdot10^{-5}$.
In order to choose the peak learning rate, we put aside random 10\% of the train data for each fold as a development set and evaluate each trained model on its corresponding development set. Finally, we choose a single peak learning rate for every model according to the 10-fold means of the development macro-averaged F1 scores. The selected peak learning rates are reported for each evaluated model.

\section{Results}

\begin{table}[t]
    \caption{Overall results.}
    \label{table:all-results}
    \centering
    \setlength{\tabcolsep}{3.4pt}
    \begin{tabular}{@{}lcccccccc@{}}
    \toprule
     & \multicolumn{2}{c}{\textbf{Morphosynt.}} & \multicolumn{2}{c}{\textbf{Morphosynt.}} & \multicolumn{2}{c}{\textbf{NER}} & \textbf{Semant.} & \textbf{Sentim.} \\
     & \multicolumn{2}{c}{\textbf{PDT3.5}} & \multicolumn{2}{c}{\textbf{UD2.3}} & \multicolumn{2}{c}{\textbf{CNEC1.1}} & \textbf{PTG} & \textbf{CFD} \\
     & \textbf{POS} & \textbf{LAS} & \textbf{XPOS} & \textbf{LAS} & \textbf{nested} & \textbf{flat} & \textbf{Avg.} & \textbf{F1} \\
    \midrule
    
    mBERT       & 98.00 & 89.74 & 97.61 & 92.34 & 86.71 & 86.45 & 90.62 & 75.43 \\
    Czert       & 98.43 & 90.68 & 98.07 & 93.13 & 85.38 & 84.69 & 90.66 & 78.52 \\
    Slavic BERT & 97.70 & 88.50 & 97.29 & 91.49 & 85.85 & 85.12 & 91.27 & 74.85 \\
    XLM-R base  & 97.62 & 88.14 & 97.29 & 91.30 & 83.25 & 82.76 & 91.55 & 79.40 \\
    XLM-R large & 98.41 & 91.27 & 98.15 & 93.49 & 87.41 & 86.86 & 92.11 & \bf 82.29 \\
    \bf RobeCzech & \bf 98.50 & \bf 91.42 & \bf 98.31 & \bf 93.77 & \bf 87.82 & \bf 87.47 & \bf 92.36 & 80.13 \\
    \midrule
    previous SoTA & 98.05 & 89.89 & 97.71 & 93.38 & 86.88 & 86.57 & 92.24 & 76.55 \\
    \bottomrule
    \end{tabular}
\end{table}

\begin{table}[p]
    \caption{Morpological tagging and lemmatization on PDT3.5.}
    \label{table:POS-lemmatization-PDT3.5-results}
    \centering
    \setlength{\tabcolsep}{3.5pt}
    \begin{tabular}{@{}lcccccc@{}}
    \toprule
      \multirow{2}{*}{\bf Model} & \multicolumn{3}{c}{\bf{Without Dictionary}} & \multicolumn{3}{c}{\textbf{With Dictionary}} \\
      & \bf POS & \bf Lemmas & \bf Both & \bf POS & \bf Lemmas & \bf Both\\
    \midrule
    mBERT & 97.86 & 98.69 & 97.21 & 98.00 & 98.96 & 97.59 \\
    Czert & 98.30 & 98.73 & 97.65 & 98.43 & 98.98 & 98.02 \\
    Slavic BERT & 97.51 & 98.58 & 96.81 & 97.70 & 98.89 & 97.27 \\
    XLM-R base & 97.43 & 98.56 & 96.76 & 97.62 & 98.85 & 97.20 \\
    XLM-R large & 98.30 & 98.76 & 97.69 & 98.41 & 98.98 & 98.01 \\
    \bf RobeCzech & \bf 98.43 & \bf 98.79 & \bf 97.83 & \bf 98.50 & \bf 99.00 & \bf 98.11 \\
    \midrule
      Mor\v{c}e (2009) \cite{spoustova09}& ---   & ---   & ---   & 95.67 & ---   & ---   \\
      MorphoDiTa (2016) \cite{ACL2014}   & ---   & ---   & ---   & 95.55 & 97.85 & 95.06 \\
      LemmaTag (2018) \cite{EMNLP2018}   & 96.90 & 98.37 & ---   & ---   & ---   & ---   \\
      UDPipe 2+mBERT+Flair \cite{StrakaTSD2019} & 97.94 & 98.75 & 97.31 & 98.05 & 98.98 & 97.65 \\
    \bottomrule
    \end{tabular}    

    \bigskip\medskip
    
    \caption{Dependency parsing on PDT3.5.}
    \label{table:parsing-PDT3.5-results}
    \centering
    \catcode`! = 13\def!{\itshape}
    \setlength{\tabcolsep}{5pt}
    \begin{tabular}{@{}lcccc@{}}
    \toprule
      \multirow{2}{*}{\bf Model} & \multirow{2}{*}{\bf UAS} & \multirow{2}{*}{\bf LAS} & \bf\itshape Joint & \bf\itshape Joint \\
      & & & \bf\itshape POS & \bf\itshape Lemmas\\
    \midrule
    mBERT & 93.01 & 89.74 & !97.62 & !98.49 \\
    Czert & 93.57 & 90.68 & !98.10 & !98.53 \\
    Slavic BERT & 92.14 & 88.50 & !97.20 & !98.29 \\
    XLM-R base & 91.80 & 88.14 & !97.22 & !98.34 \\
    XLM-R large & 94.07 & 91.27 & !98.12 & !98.54 \\
    \bf RobeCzech & \bf 94.14 & \bf 91.42 & \bf !98.28 & \bf !98.62 \\
    \midrule
      UDPipe 2+mBERT+Flair \cite{StrakaTSD2019} & 93.07 & 89.89 & !97.72 & !98.51 \\
    \bottomrule
    \end{tabular}

    \bigskip\medskip
    
    \caption{Morphosyntactic analysis on UD 2.3. Models marked $^f$ are fine-tuned, otherwise with frozen embeddings.}
    \label{table:UD2.3-results}
    \centering
    \setlength{\tabcolsep}{1.2pt}
    \begin{tabular}{@{}lcccccccc@{}}
    \toprule
      \bf Model & \bf UPOS & \bf XPOS & \bf UFeats & \bf Lemmas & \bf UAS & \bf LAS & \bf MLAS & \bf BLEX \\
    \midrule
    mBERT & 99.31 & 97.61 & 97.55 & 99.06 & 94.27 & 92.34 & 87.75 & 89.91 \\
    Czert & 99.32 & 98.07 & 98.05 & 99.09 & 94.75 & 93.13 & 89.19 & 90.92 \\
    Slavic BERT & 99.22 & 97.29 & 97.22 & 98.99 & 93.53 & 91.49 & 86.37 & 88.79 \\
    XLM-R base & 99.18 & 97.29 & 97.24 & 99.02 & 93.32 & 91.30 & 86.18 & 88.62 \\
    XLM-R large & 99.36 & 98.15 & 98.10 & 99.17 & 95.15 & 93.49 & 89.64 & 91.40 \\
    \bf RobeCzech & \bf 99.36 & \bf 98.31 & \bf 98.28 & \bf 99.18 & \bf 95.36 & \bf 93.77 & \bf 90.18 & \bf 91.82 \\
    \midrule
      \makecell[l]{UDPipe 2\\~~+mBERT+Flair \cite{StrakaTSD2019}} & 99.34 & 97.71 & 97.67 & 99.12 & 94.43 & 92.56 & 88.09 & 90.22 \\
      UDify$^f$ \cite{udify}                    & 99.24 & ---   & 94.77 & 98.93 & 95.07 & 93.38 & ---   & ---   \\
      Czert$^f$ \cite{sido2021czert}            & 99.30 & ---   & ---   & ---   & ---   & ---   & ---   & ---   \\
    \bottomrule
    \end{tabular}
\end{table}

\begin{table}[p]
\caption{Named entity recognition F1 scores (3 runs average) in comparison with previous reports. Models marked $^f$ are fine-tuned, otherwise with frozen embeddings.}
\label{table:ner-results}
\centering
\setlength{\tabcolsep}{4.2pt}
\begin{tabular}{@{}lcccc@{}}
\toprule
 
 \textbf{Model} & \textbf{CNEC1.1} & \textbf{CNEC2.0} & \makecell{\textbf{CoNLL} \\ \textbf{CNEC1.1}} & \makecell{\textbf{CoNLL}\\\textbf{CNEC2.0}} \\
\midrule
mBERT & 86.71 & 84.21 & 86.45 & 87.04 \\
Czert & 85.38 & 82.84 & 84.69 & 85.33 \\
Slavic BERT & 85.85 & 82.71 & 85.12 & 85.28 \\
XLM-R base & 83.25 & 80.33 & 82.76 & 82.85 \\
XLM-R large & 87.41 & 84.46 & 86.86 & 87.06 \\
\bf RobeCzech & \bf 87.82 & \bf 85.51 & \bf 87.47 & \textbf{87.49} \\
\midrule
seq2seq+mBERT \cite{StrakaTSD2019,StrakovaACL2019} & 86.73 & 84.66 & --- & --- \\
seq2seq+mBERT+Flair \cite{StrakaTSD2019,StrakovaACL2019} & 86.88 & 84.27 & --- & --- \\
LSTM-CRF, LDA \cite{KonopikPrazak2018} & --- & --- & 81.77 & --- \\
LSTM-CRF \cite{StrakovaSaS2019} & 83.15 & --- & 83.27 & 84.22 \\
LSTM-CRF+BERT \cite{Muller2020} & --- & --- & --- & 86.39 \\
Czert${^f}$ \cite{sido2021czert} & --- & --- & 86.27 & --- \\
mBERT${^f}$ \cite{devlin2018bert}, by \cite{sido2021czert} & --- & --- & 86.23 & --- \\
Slavic BERT${^f}$ \cite{SlavicBERT}, by \cite{sido2021czert} & --- & --- & 86.57 & --- \\
\bottomrule
\end{tabular}

    \bigskip

\caption{Semantic parsing F1 scores on Prague Tectogrammatical Graphs.}
\label{table:ptg-results}
\resizebox{\textwidth}{!}{%
\setlength{\tabcolsep}{3pt}
\begin{tabular}{@{}lcccccc@{}}
\toprule
\textbf{Model} & \textbf{Labels} & \textbf{Properties} & \textbf{Anchors} & \textbf{Edges} & \textbf{Attributes} & \textbf{Average} \\ \midrule
mBERT & 95.72 & 92.60 & 97.20 & 80.77 & 72.83 & 90.62 \\
Czert & 95.72 & 92.69 & 97.23 & 80.91 & 72.37 & 90.66 \\
Slavic BERT & 95.92 & 92.91 & 97.51 & 82.48 & 75.08 & 91.27 \\
XLM-R base & 96.09 & 93.12 & 97.60 & 83.03 & 76.16 & 91.55 \\
XLM-R large & 96.42 & 93.31 & 97.92 & 84.46 & 77.89 & 92.11 \\
RobeCzech frozen & 95.85 & 92.76 & 97.41 & 82.60 & 74.95 & 91.23 \\
\textbf{RobeCzech} & \textbf{96.57} & \textbf{93.58} & \textbf{97.97} & \textbf{84.92} & \textbf{78.29} & \textbf{92.36} \\
\midrule
HUJI-KU+mBERT \cite{arviv-etal-2020-huji} & --- & 72.44 & 72.10 & 44.91 & --- & 58.49 \\
HIT-SCIR+mBERT \cite{dou-etal-2020-hit} & 84.14 & 79.01 & 92.34 & 64.96 & 47.68 & 77.93 \\ 
Hitachi+mBERT \cite{ozaki-etal-2020-hitachi} & 87.69 & 91.48 & 93.99 & 76.90 & 66.07 & 87.35 \\
ÚFAL+XLM-R large \cite{samuel-straka-2020-ufal} & 96.23 & 93.56 & 97.86 & 84.61 & 78.62 & 92.24 \\
\bottomrule
\end{tabular}%
}

    \bigskip

    \caption{Sentiment analysis 10-fold macro F1 scores on Czech Facebook dataset.}
    \label{table:sentiment-analysis-results}
    \centering
    \setlength{\tabcolsep}{7pt}
    \begin{tabular}{@{}lccc@{}}
    \toprule
      \bf Model & \bf 10-fold Macro F1 & \bf 10-fold Std & \bf Chosen LR \\
    \midrule
      mBERT & 75.43 & $\pm$1.38 & $5\cdot10^{-5}$ \\
      Czert & 78.52 & $\pm$1.16 & $2\cdot10^{-5}$ \\
      Slavic BERT & 74.85 & $\pm$1.27 & $5\cdot10^{-5}$ \\
      XLM-R base & 79.40 & $\pm$1.07 & $1\cdot10^{-5}$ \\
      XLM-R large & \bf 82.29 & $\pm$1.19 & $1\cdot10^{-5}$ \\
      \bf RobeCzech & 80.13 & $\pm$1.21 & $3\cdot10^{-5}$ \\
    \midrule
      Czert \cite{sido2021czert} & 76.55 & --- & $3\cdot10^{-6}$ \\
      MaxEnt \cite{habernal-etal-2013-sentiment} & 69.4\hphantom{0} & --- & --- \\
    \bottomrule
    \end{tabular}
\end{table}

Table~\ref{table:all-results} summarizes the overall results of all considered language representation models in all evaluated tasks. RobeCzech improves over current state of the art in all five evaluated NLP tasks, and at the same time, clearly outperforms current multilingual and Czech-trained contextualized language representation models, being surpassed only in one of the five tasks by a model 4 times its size (XLM-RoBERTa large \cite{XLM-R}, Table~\ref{table:all-results}). Notably, RobeCzech reaches 25\% error reduction in POS tagging both on PDT 3.5 and UD 2.3, and 15\% error reduction in dependency parsing on PDT 3.5, significantly improving performance of Czech morphosyntactic analysis.

Furthermore, for each of the evaluated tasks, we show the detailed results in Tables~\ref{table:POS-lemmatization-PDT3.5-results}, \ref{table:parsing-PDT3.5-results}, \ref{table:UD2.3-results}, \ref{table:ner-results}, \ref{table:ptg-results} and \ref{table:sentiment-analysis-results}.

The results demonstrate that the large variant of XLM-RoBERTa reaches considerably better results compared to base size of other multilingual models. Yet, RobeCzech still surpasses it on four tasks, most notably in the frozen scenario. We hypothesize that in the frozen scenario the larger model cannot capitalize on its superior capacity, compared to for example sentiment analysis, where its capacity proves determining.

\section{Conclusion}

We introduced RobeCzech, a Czech contextualized language representation model based on RoBERTa. We described the training process and we evaluated RobeCzech in comparison with, to our best knowledge, currently known multilingual and Czech-trained contextualized language representation models. We show that RobeCzech considerably improves over state of the art in all five evaluated NLP tasks. Notably, it yields 25\% error reduction in POS tagging both on PDT~3.5 and UD~2.3 and 15\% error reduction in dependency parsing on PDT 3.5. We publish RobeCzech publicly at \lindaturl and \hfurl.

\section*{Acknowledgements}

\looseness-1
The research described herein has been supported by the Czech Science
Foundation grant No. GX20-16819X, Mellon Foundation Award No. G-1901-06505 and
by the Grant Agency of Charles University 578218. The resources used have
been provided by the LINDAT/CLARIAH-CZ Research Infrastructure, project No.
LM2018101 of the Ministry of Education, Youth and Sports of the Czech Republic.



%
\pdfoutput=1
%

%
%

\bibliographystyle{splncs04}
\bibliography{robeczech}

\end{document}